\documentclass[conference]{IEEEtran}
\IEEEoverridecommandlockouts

\usepackage{amsmath,amssymb,amsfonts}
\usepackage[backend=bibtex, sorting=none]{biblatex}
\usepackage{graphicx}
\usepackage{subcaption} % in preamble
\usepackage{textcomp}
\usepackage{xcolor}
\usepackage{booktabs} 
\usepackage{bm}
\usepackage{amsthm}
\usepackage{algorithm}
\usepackage{algorithmicx}
\usepackage{algpseudocode}
\usepackage{tikz}
\usepackage{tabularx}
\usepackage{soul}  % highlights
\usetikzlibrary{positioning}
\usetikzlibrary{shapes.geometric, arrows.meta, positioning, calc}
\usetikzlibrary{shapes.geometric, arrows.meta, positioning, calc}
\theoremstyle{plain}

\theoremstyle{definition}

\theoremstyle{remark}

\def\BibTeX{{\rm B\kern-.05em{\sc i\kern-.025em b}\kern-.08em
    T\kern-.1667em\lower.7ex\hbox{E}\kern-.125emX}}
\usepackage[acronym]{glossaries}
\newacronym{gnn}{GNN}{Graph Neural Network}
\newacronym{gat}{GAT}{Graph Attention Network}
\newacronym{nn}{NN}{Neural Network}
\newacronym{cnn}{CNN}{Convolutional Neural Network}
\newacronym{mpnn}{MPNN}{Message Passing Neural Network}
\newacronym{ae}{AE}{Autoencoder}
\newacronym{vae}{VAE}{Variational Autoencoder}
\newacronym{vgae}{VGAE}{Variational Graph Autoencoder}
\newacronym{gru}{GRU}{Gated Recurrent Unit}
\newacronym{mse}{MSE}{Mean Squared Error}
\newacronym{pf}{PF}{Power Flow}
\newacronym{nr}{NR}{Newton--Raphson}

\begin{document}

\title{Power Flow Feasibility Assessment Using Variational Graph Autoencoders\vspace{-3mm}\\

\thanks{The research has been funded by projects Daedalos(Horizon Europe research grant agreement No 101172829), GRAPHS4SEC (grant PCI2023-145974-2 funded by MICIU/AEI/10.13039/501100011033) and BLOSSOMS (grant PID2024-158530OB-I00, by MI-CIU/AEI/10.13039/501100011033/ and ERDF/EU). The work of P.Barlet, O.Gomis-Bellmunt and E.Prieto-Araujo was supported by the Agència de Gestió d’Ajuts Universitaris i de Recerca (AGAUR) through the ICREA Acadèmia programme, and by the Departament de Recerca i Universitats of the Generalitat de Catalunya. E. Prieto-Araujo is a member of the Serra Húnter Programme.}
}

\author{\IEEEauthorblockN{Ferran Bohigas-Daranas}
\IEEEauthorblockA{\textit{CITCEA} \\
\textit{UPC}\\
Barcelona, Spain \\
0009-0001-6477-6591}
\and
\IEEEauthorblockN{Hamid Latif-Martinez}
\IEEEauthorblockA{\textit{BNN} \\
\textit{UPC}\\
Barcelona,Spain \\
0000-0001-6006-3175}
\and
\IEEEauthorblockN{Eduardo Prieto-Araujo}
\IEEEauthorblockA{\textit{CITCEA} \\
\textit{UPC}\\
Barcelona, Spain \\
0000-0003-4349-5923}
\and
\IEEEauthorblockN{Pere Barlet-Ros}
\IEEEauthorblockA{\textit{BNN} \\
\textit{UPC, Hypergraph}\\
Barcelona,Spain \\
0000-0001-7837-0886}
\and
\IEEEauthorblockN{Oriol Gomis-Bellmunt}
\IEEEauthorblockA{\textit{CITCEA} \\
\textit{UPC}\\
Barcelona, Spain \\
0000-0002-9507-8278}
}

\maketitle

\begin{abstract}

Data-driven methods, including graph neural networks, have been studied for accelerating power flow calculations in recent years, but very little attention has been paid to the solution feasibility, which can be obtained by traditional solvers. This paper presents a \gls{vgae} that detects the power flow solution feasibility, using the IEEE 118-bus case, to assess the validity of the solutions provided by AI-driven solvers.

\end{abstract}
 
% ═══════════════════════════════════════════════════════════════
\section{Introduction}
% ═══════════════════════════════════════════════════════════════

Steady-state analysis of electric power networks relies on the solution of a system of nonlinear algebraic equations known as the \gls{pf} equations, by using a set of nonlinear equality constraints encoding Kirchhoff's laws.
 
In practice, the PF problem is solved by iterative algorithms, such as~\gls{nr} \cite{tinney_power_1967}, whose convergence is not guaranteed, even when mathematical research has been carried out to improve the results \cite{klump_techniques_2000}, and new methods have been proposed, such as holomorphic embedding (HELM) \cite{Trias2012} for power flow calculation. Power system operators frequently encounter cases in which a solver fails to converge, and the fundamental question arises: does the problem admit no solution (infeasibility), or did the algorithm simply fail to find one (non-convergence)? 

The terms \emph{feasibility} and \emph{convergence} are frequently confused in the power systems literature, yet they refer to fundamentally distinct mathematical properties. Feasibility is a property of a \emph{problem}: it concerns the existence of a solution satisfying all governing equations and constraints. Convergence is a property of an \emph{algorithm}: it concerns whether an iterative procedure finds that solution.
 
This distinction carries critical operational consequences. An infeasible PF indicates that the system has surpassed its voltage stability margin. Convergence failure, by contrast, may simply require better initialization, rescaling, or a more robust algorithm. Misdiagnosis leads to either unnecessary load curtailment or, more dangerously, to false confidence in a solution that was never actually found.

This problem can be still more dangerous when using AI or data-driven methods, which provide a regression solution that does not consider the feasibility. Prior work has focused solely on the regression solution \cite{donon_neural_2020}\cite{fatah_integrating_2021}, neglecting problem feasibility; only Li et al. \cite{li_convergence_2024} have proposed a~\gls{gat} architecture to predict power flow convergence using supervised learning.

This paper proposes a fast and reliable method to assess the feasibility of the PF problem by using a \gls{vgae} architecture based on~\glspl{mpnn}.

The remainder of this paper is organized as follows. Section~\ref{sec:pf} presents the PF problem, its feasibility theory, and the classification of convergence failures. Section~\ref{sec:gnn_ae} details what~\glspl{gnn} are and the autoencoder-related architectures. Section~\ref{sec:architecture} details our implementation to assess the PF feasibility, while Section~\ref{sec:Results} shows the results obtained on our tests. Section~\ref{sec:Conclusions} summarizes the work.
 
% ═══════════════════════════════════════════════════════════════
\section{Power Flow: Feasibility and Convergence}
\label{sec:pf}
% ═══════════════════════════════════════════════════════════════
 
\subsection{Problem Formulation}
 
Let $\mathcal{N} = \{0,1,\ldots,n\}$ denote the set of buses, with bus $0$ designated as the slack bus. The complex admittance matrix $\mathbf{Y}_\text{bus} \in \mathbb{C}^{(n+1)\times(n+1)}$ encodes the network topology and line parameters. The PF equations express the complex power injection at each bus $i \in \mathcal{N} \setminus \{0\}$:
 
\begin{equation}
  S_i = P_i + jQ_i = V_i \sum_{k \in \mathcal{N}} Y_{ik}^{*} V_k^{*},
  \label{eq:pf_complex}
\end{equation}
 
\noindent where $V_i = |V_i|e^{j\theta_i}$ is the complex bus voltage. Separating real and imaginary parts yields the standard polar-form PF equations:
 
\begin{align}
  P_i &= |V_i| \sum_{k \in \mathcal{N}} |V_k|\bigl(G_{ik}\cos\theta_{ik} + B_{ik}\sin\theta_{ik}\bigr), \label{eq:pf_P}\\
  Q_i &= |V_i| \sum_{k \in \mathcal{N}} |V_k|\bigl(G_{ik}\sin\theta_{ik} - B_{ik}\cos\theta_{ik}\bigr), \label{eq:pf_Q}
\end{align}
 
\noindent where $\theta_{ik} = \theta_i - \theta_k$, and $G_{ik}$ (conductance), $B_{ik}$ (susceptance) are the real and imaginary parts of $Y_{ik}$, respectively. This constitutes a system of $2(n)$ real equations in $2(n)$ unknowns $(\bm{\theta}, |\mathbf{V}|)$, compactly written as:
 
\begin{equation}
  \mathbf{f}(\mathbf{x}) = \mathbf{0}, \quad \mathbf{x} = \bigl[\bm{\theta};\, |\mathbf{V}|\bigr] \in \mathbb{R}^{2n}.
  \label{eq:pf_compact}
\end{equation}
 
\subsection{Saddle-Node Bifurcation and the Feasibility Boundary}

The PF problem~\eqref{eq:pf_compact} is \emph{feasible} if there exists at least one $\mathbf{x}^* \in \mathbb{R}^{2n}$ such that $\mathbf{f}(\mathbf{x}^*) = \mathbf{0}$.
The set of all solutions forms a smooth \emph{solution manifold} in the joint space of injections and voltages. 

For a canonical two-bus system with a slack bus voltage $V_1 = 1.0$\,pu, line impedance $Z = R + jX$, and load $P + jQ$, substituting~\eqref{eq:pf_P}--\eqref{eq:pf_Q} into the Kirchhoff relations and eliminating angles yields a quadratic equation in $V_2^2$:
 
\begin{equation}
  V_2^4 - \bigl[V_1^2 - 2(RP + XQ)\bigr] V_2^2 + (R^2+X^2)(P^2+Q^2) = 0
  \label{eq:two_bus_quadratic}
\end{equation}
 
The discriminant ($\Delta$) of~\eqref{eq:two_bus_quadratic} determines feasibility:
 
\begin{equation}
  \Delta = \bigl[V_1^2 - 2(RP+XQ)\bigr]^2 - 4(R^2+X^2)(P^2+Q^2)
  \label{eq:discriminant}
\end{equation}
 
\begin{figure}
    \centering
    \includegraphics[width=1\linewidth]{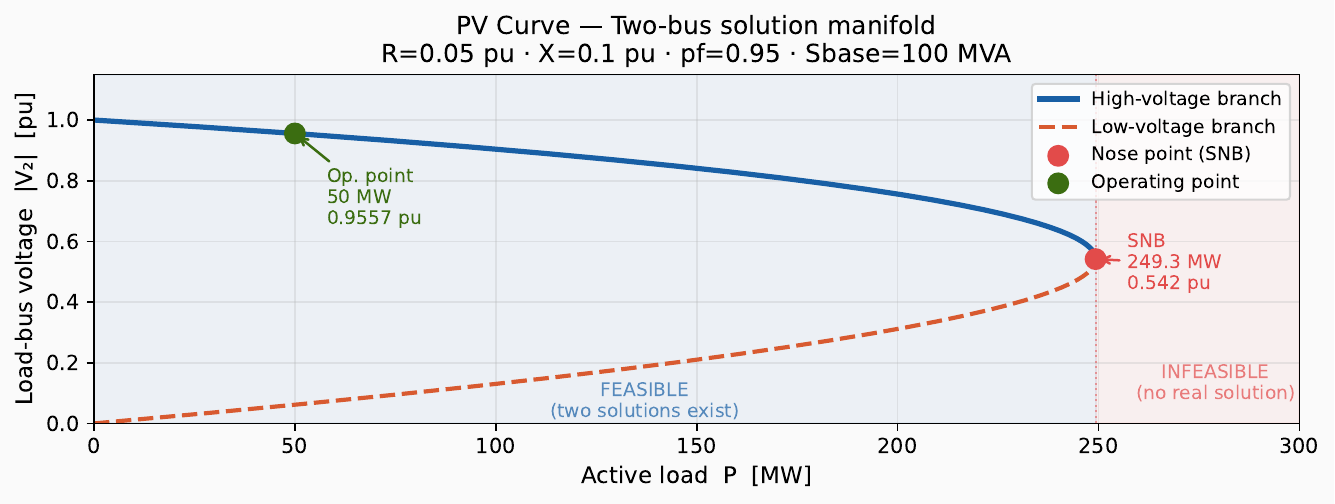}
    \caption{Nose or saddle-node bifurcation (SNB) point }
    \label{fig:SNB}
\end{figure}

When $\Delta > 0$, two real solutions exist: the high-voltage (stable) and low-voltage (unstable) branches. When $\Delta = 0$, the two branches merge at the \emph{nose point} or \emph{saddle-node bifurcation} (SNB)  (Figure \ref{fig:SNB}). Beyond this point, when $\Delta < 0$, no real solution exists, so the system is infeasible. %The \emph{SNB} can be also defined as the point at which the Jacobian of the PF equations becomes singular \textbf{REMOVE to ARRIVE TO 5p}:
 
%\begin{equation}
%  \mathbf{J}(\mathbf{x}) \triangleq \frac{\partial \mathbf{f}}{\partial \mathbf{x}}, \qquad \det\bigl(\mathbf{J}(\mathbf{x}^*)\bigr) = 0.
%  \label{eq:snb}
%\end{equation}
 
\subsection{Newton--Raphson for Power Flow solution}
The most widely used algorithm to solve the PF equations is the~\gls{nr} iterative algorithm.
In practice, NR converges in 3--7 iterations \cite{tinney_power_1967} from a flat start for well-conditioned systems. Convergence failure occurs, however, in several practically relevant situations that are \emph{distinct} from infeasibility:
 
\begin{enumerate}
  \item Initialization outside the solution range. 
  \item Near-singular Jacobian. 
  \item Generator reactive limit switching. 
  \item Erroneous network data. 
\end{enumerate}
 
Table~\ref{tab:pf_matrix} summarizes the four possible combinations and their operational interpretations.
 
\begin{table}[!t]
  \renewcommand{\arraystretch}{1.3}
  \caption{Feasibility vs. Convergence in PF}
  \label{tab:pf_matrix}
  \centering
  \small
  \begin{tabularx}{\columnwidth}{@{}l XX@{}}
    \toprule
    & \textbf{Feasible} & \textbf{Infeasible} \\
    \midrule
    \textbf{Converges}  & Correct solution found & False convergence / spurious solution \\
    \textbf{Diverges}   & Algorithmic failure & Correct detection of infeasibility \\
    \bottomrule
  \end{tabularx}
\end{table}

Divergence is evidence about the \emph{algorithm}, not the physics. Only after ruling out data errors, poor initialization, and numerical ill-conditioning can divergence be attributed to genuine infeasibility, and even then, analytical confirmation via continuation power flow (CPF) \cite{ajjarapu_continuation_1992} or HELM methods, which provide complete certainty about feasibility, is advisable.
\section{Graph Neural Network Autoencoders}
\label{sec:gnn_ae}
 
\subsection{Graphs as a Data Structure}
 
A power network is naturally represented as a graph $\mathcal{G} = (\mathcal{V}, \mathcal{E}, \mathbf{X}, \mathbf{E})$, where $\mathcal{V}$ is the set of $N$ nodes (buses), $\mathcal{E} \subseteq \mathcal{V} \times \mathcal{V}$ is the set of edges (branches), $\mathbf{X} \in \mathbb{R}^{N \times d_v}$ is the node feature matrix (each row $\mathbf{x}_i$ holds the $d_v$ features of node $i$), and $\mathbf{E} \in \mathbb{R}^{|\mathcal{E}| \times d_e}$ collects the $d_e$ edge features. The topology is encoded by the adjacency matrix $\mathbf{A} \in \{0,1\}^{N \times N}$, with $A_{ij} = 1$ if $(i,j) \in \mathcal{E}$.
% Standard neural networks are unable to operate on this structure because they assume a fixed-size Euclidean input; Graph Neural Networks (GNNs) are designed precisely to lift this limitation.

While~\glspl{cnn} are designed for Euclidean-structured data, they cannot operate on irregular structures such as graphs. ~\glspl{gnn} \cite{scarselli2008graph} are capable of operating directly on these data, learning both local and global patterns inherent in the graph topology. 
 
\subsection{Graph Neural Networks}
 
% A~\gls{gnn} learns a function $\phi : \mathcal{G} \to \mathcal{Z}$ that maps a graph to a set of node embeddings $\mathbf{Z} \in \mathbb{R}^{N \times d_z}$ by iteratively aggregating information from local neighbourhood. At each layer $\ell$, the embedding of node $i$ is updated by a two-step \emph{message-passing} scheme:
 
% \begin{align}
%   \mathbf{m}_i^{(\ell)} &= \text{AGG}\!\left(
%       \left\{ \psi\!\left(\mathbf{h}_i^{(\ell-1)},\,
%                           \mathbf{h}_j^{(\ell-1)},\,
%                           \mathbf{e}_{ij}\right)
%               : j \in \mathcal{N}(i) \right\} \right), \label{eq:mp_agg}\\
%   \mathbf{h}_i^{(\ell)} &= \text{UPDATE}\!\left(
%       \mathbf{h}_i^{(\ell-1)},\, \mathbf{m}_i^{(\ell)} \right), \label{eq:mp_upd}
% \end{align}
 
% \noindent where $\mathbf{h}_i^{(0)} = \mathbf{x}_i$, $\mathcal{N}(i)$ is the set of neighbours of $i$, $\mathbf{e}_{ij}$ are the edge features, and $\psi$, AGG, UPDATE are learnable or fixed functions (e.g.\ multi-layer perceptrons and sum/mean aggregators).  

% After $L$ layers, $\mathbf{h}_i^{(L)}$ encodes the structural and feature information within an $L$-hop neighbourhood of node $i$.

\glspl{gnn} are neural architectures designed to operate on graph-structured data by learning node representations that incorporate both feature and structural information. In this work, we adopt the~\gls{mpnn} formulation~\cite{gilmer2017neural}, which represents the most generic~\gls{gnn} framework.

Given a graph $\mathcal{G}$, each node $i$ is initialized with a feature vector $\mathbf{h}_i^{(0)} = \mathbf{x}_i$. The model then performs $T$ iterations of message passing, where node representations are iteratively refined by exchanging information with their neighbours:
\begin{align}
  \mathbf{m}_i^{(t)} &= \text{AGG}\!\left(
      \left\{ \psi\!\left(\mathbf{h}_i^{(t-1)},\,
                          \mathbf{h}_j^{(t-1)},\,
                          \mathbf{e}_{ij}\right)
              : j \in \mathcal{N}(i) \right\} \right), \\
  \mathbf{h}_i^{(t)} &= \text{UPDATE}\!\left(
      \mathbf{h}_i^{(t-1)},\, \mathbf{m}_i^{(t)} \right).
\end{align}

After $T$ iterations, the embeddings $\mathbf{h}_i^{(T)}$ encode information from the $T$-hop neighbourhood of each node, capturing both local attributes and structural context. Then, these learned representations can be used as inputs to downstream models for any desired task (such as classification). In this work, we leverage them within an autoencoder framework to learn compact representations of the graph.

\subsection{Graph Autoencoders for Learning Feasible Network Structures}

\glspl{ae} learn a compressed latent representation of data and reconstruct the original input from it. At a high level, an~\gls{ae} consists of two core components: the encoder, which compresses the input data into a lower-dimensional representation, and the decoder, which attempts to reconstruct the original input based on this compressed embedding. The model is trained to reconstruct the compressed representation so that it is as close as possible to the original input.

Formally, given an input $\mathbf{x}$, the encoder produces a latent representation $\mathbf{z} = f_\phi(\mathbf{x})$, and the decoder reconstructs the input as $\hat{\mathbf{x}} = g_\theta(\mathbf{z})$. Training minimizes the reconstruction loss of the form :
\begin{equation}
\mathcal{L} = \mathbb{E}_{\mathbf{x} \sim p_\text{data}} \left[ \mathcal{D}\left(\mathbf{x}, \hat{\mathbf{x}}\right) \right],
\end{equation}
where $\mathcal{D}$ measures the difference between the original input and the reconstructed output.

The key idea is that, by compressing the input into a latent space, the model is forced to retain only the most important features and patterns present in the input. As a result, the autoencoder learns to reconstruct inputs that are similar to those seen during training, while struggling to reconstruct inputs that differ from the training distribution.

This principle can be naturally extended to graph-structured data by using~\glspl{gnn}. In this setting, the input is a graph $\mathcal{G} = (\mathbf{A}, \mathbf{X})$, where $\mathbf{A}$ represents the topology and $\mathbf{X}$ the node features. The encoder is implemented as a~\gls{gnn} that produces node-level embeddings:
\begin{equation}
\mathbf{Z} = f_\phi(\mathbf{A}, \mathbf{X}),
\end{equation}
where each embedding captures both its local features and the structure of its neighbourhood. Then, the decoder attempts to reconstruct the original graph from these embeddings.

In this paper, we apply this principle to learn representations of \textit{feasible network configurations}. By training the model exclusively on regular (i.e., physically or operationally valid) network states, the autoencoder learns to accurately reconstruct input graphs that follow the underlying constraints of the system. However, when fed a graph representing an anomalous or infeasible network, the model is unable to properly reconstruct it, leading to a significantly higher reconstruction error.

In this way, we train the model to capture what constitutes a valid network configuration and flag deviations from this learned structure. In other words, the autoencoder acts as a filter that recognizes whether a given sample resembles the set of feasible networks it has seen in training or not.

\section{Architecture Description}
\label{sec:architecture}

Building on the graph autoencoder framework introduced in the previous section, we now describe the specific architecture used to model feasible network configurations. 

Our model follows a~\gls{vgae} design, where the encoder is implemented as a~\gls{mpnn} and the decoder reconstructs node features from the learned latent representations.

\subsection{Message-Passing Encoder}

The encoder is based on the~\gls{mpnn} formulation, where node representations are iteratively refined by exchanging messages with their neighbors. Starting from the input node features, each node updates its hidden representation over $T$ message-passing iterations.

At each iteration, messages are computed along edges using a learnable function that depends on: $(i)$ the source node representation, $(ii)$ the destination node representation, and $(iii)$ the edge attributes.

These messages are then aggregated at each node to produce a summary of its neighbourhood. To weight neighbor contributions differently, we incorporate an attention mechanism. Each incoming message is assigned a learnable coefficient, normalized to sum to one using a softmax function. This enables each node to prioritize the most relevant connections when updating its hidden representation.

Before updating the node state, a residual connection is added between the previous node representation and the aggregated neighbourhood message. This allows the model to preserve the current node state while learning only the message-based correction, which improves optimization stability and facilitates information flow across message-passing iterations~\cite{he2016deep}.

After $T$ iterations, each node embedding encodes both its local features and the structure of its surrounding neighbourhood. These representations are then used to parameterize a Gaussian distribution per node, producing a mean $\boldsymbol{\mu}_i$ and variance $\boldsymbol{\sigma}_i^2$. This formulation follows the~\gls{vgae} framework~\cite{kipf2016variational}, where latent variables are sampled from the learned distributions. To enable backpropagation through the sampling process, the reparameterization trick is applied:
\begin{equation}
\mathbf{z}_i = \boldsymbol{\mu}_i + \boldsymbol{\sigma}_i \odot \boldsymbol{\epsilon}, \quad \boldsymbol{\epsilon} \sim \mathcal{N}(0, I)
\end{equation}

\subsection{Decoder and Reconstruction Objective}

The decoder maps each node embedding to a reconstruction of the original node features. Then, we use the error between the reconstruction and the input to train the model.

Since the model is trained exclusively on valid network states, it learns to accurately reconstruct feasible configurations. In this way, when presented with an anomalous configuration, the reconstruction error (\ref{eq:VGAElosseq}) increases significantly, providing a straightforward anomaly score.

\begin{equation}
\mathcal{L} = \mathcal{L}_{\text{recon}} + \beta \cdot D_{\mathrm{KL}}\!\left(\mathcal{N}(\mu_i, \sigma_i^2) \,\|\, \mathcal{N}(\mathbf{0}, \mathbf{I})\right)
\label{eq:VGAElosseq}\end{equation}

\section{Experimental Evaluation}
\label{sec:Results}
\subsection{Dataset Generation}
The dataset generation is a key element of the pipeline, as it determines what the model considers feasible. For that reason, it is key to create a dataset that covers different infeasibility causes, including load variations (since high demand can cause voltage collapse), topological changes such as line and bus removal, and constraint modifications such as changes to active and reactive power limits. An insufficiently diverse dataset can lead to trivially simple classifications, such as considering only the amount of power, the only infeasibility cause. 

To train and test our proposed~\gls{vgae} architecture, we have created three data sets, based on the IEEE 118-bus case, by modifying some network parameters randomly:
\begin{itemize}
    \item Stressed Continuous dataset (SC): where the load power is randomly modified from 20\% to 400\% of their nominal value, while 80\% of the samples have topology changes (lines or buses removed).
    \item Relaxed Continuous dataset (RC): where the load power is modified between 50\% to 250\%, without topology changes, to force the~\gls{vgae} to focus only on the load power as criteria.
        \item Stressed Separated dataset (SS): where the load power is randomly modified from 20\% to 150\% and 300\% to 400\% of their nominal value, clearly separating overload cases, while 80\% of the samples have topology changes (lines or buses removed).
\end{itemize}
The final test set comprises 1,500 samples, constructed by combining equal proportions from the three aforementioned scenarios, including positive and negative cases, while two training sets have been used, each of them with 500 feasible samples, one for the Stressed Continuous and the other for the Stressed Separated case.

\subsection{Training Procedure}

The model is trained as a~\gls{vgae} using only feasible (regular) grid samples. The objective function combines a reconstruction loss and a regularization term on the latent space. The reconstruction loss is defined as the~\gls{mse} between the original node features $\mathbf{x}$ and their reconstruction $\hat{\mathbf{x}}$. 

% Additionally, a Kullback--Leibler (KL) divergence term enforces a Gaussian prior over the latent representations:
% \begin{equation}
% \mathcal{L} =
% \underbrace{\text{MSE}(\hat{\mathbf{x}}, \mathbf{x})}_{\text{reconstruction}} +
% \beta \cdot
% \underbrace{\text{KL}\big(q(\mathbf{z}|\mathbf{x}) \,\|\, \mathcal{N}(0, I)\big)}_{\text{regularization}}
% \end{equation}
% where $\beta = 10^{-3}$ controls the contribution of the latent regularization.

The model is trained for 50 epochs using the Adam optimizer, the learning rate was initially set to $10^{-3}$, and decayed by a factor of 0.5 at epoch 30.

Importantly, the model is \textit{never exposed to anomalous (non-feasible) graphs during training}, ensuring that anomaly detection relies purely on reconstruction behavior.

\subsection{Performance Before Training}

\begin{figure}[t]
    \centering
    \subcaptionbox{Before training.\label{fig:histogram_total_pretraining}}
        {\includegraphics[width=0.48\linewidth]{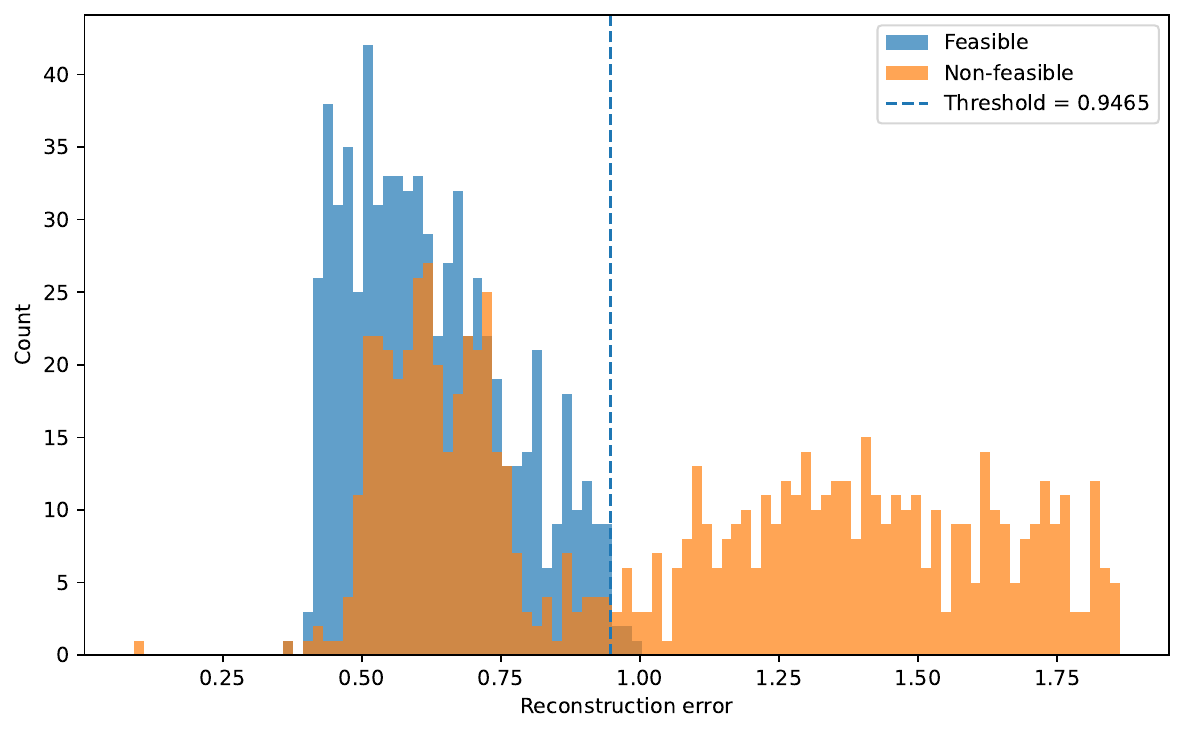}}
    \hfill
    \subcaptionbox{After training.\label{fig:histogram_total_posttraining}}
        {\includegraphics[width=0.48\linewidth]{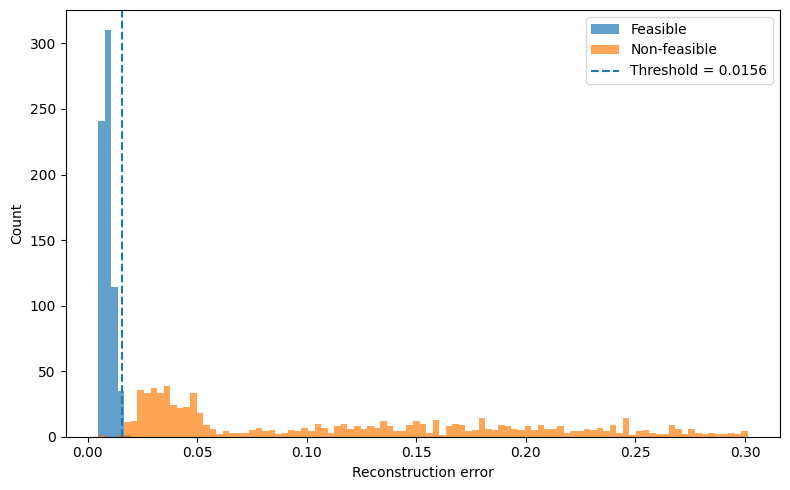}}
    \caption{Score histogram before and after training}
    \label{fig:histograms}
\end{figure}

Before training, the model parameters are randomly initialized and do not encode any knowledge about feasible grid configurations. As a result, the reconstruction error is similar for both feasible and anomalous samples, with no clear separation between the two classes. Fig.~\ref{fig:histogram_total_pretraining} shows the distribution of reconstruction errors prior to training. Both feasible and non-feasible configurations exhibit a substantial overlap, with no clear separation, indicating that the reconstruction error does not provide a meaningful signal for anomaly detection at this stage.

\subsection{Reconstruction Behavior Across Classes}

\begin{figure}[t]
    \centering
    %\subcaptionbox{Training Loss.\label{fig:training_loss}}
    %    {\includegraphics[width=0.4\linewidth]{images/loss_training.png}}
    %\hfill
    %\subcaptionbox{Class differentiation during training.\label{fig:differentiation}}
        {\includegraphics[width=0.7\linewidth]{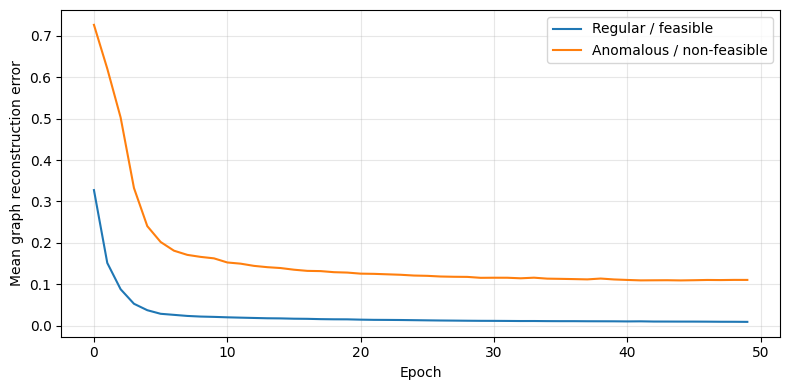}}
    \caption{Class differentiation during raining.\label{fig:differentiation}}
    %\label{fig:training}
\end{figure}

The evolution of the reconstruction error for both classes during training is shown in Fig.~\ref{fig:differentiation}. As expected, the reconstruction error for feasible graphs decreases steadily during training, showing that the model is learning to encode and reconstruct regular grid configurations. The reconstruction error of anomalous configurations also decreases over time, which reflects the way the dataset is constructed: anomalous samples are not arbitrarily generated, but rather remain structurally close to feasible configurations. Consequently, the model is able to partially generalize and reconstruct some aspects of these graphs, even though without the same precision as with the regular graphs. 

This property is desirable because it better reflects realistic scenarios, where the anomalous grid states may still share many characteristics with valid ones.

\subsection{Detection Performance}

Anomaly detection is performed by thresholding the graph reconstruction error. Specifically, the threshold is defined as the 99th percentile of the reconstruction error on the validation dataset. Since the validation dataset only contains feasible configurations, the expected reconstruction error is low and, thus, selecting a high percentile ensures that most regular samples fall below the threshold. This percentile-based approach provides an unbiased way to define the decision boundary, reducing the need to fine-tune the threshold for the specific test scenario. 

\begin{table}[!t]
\renewcommand{\arraystretch}{1.3}
\caption{Classification naming}
\label{tab:confusion_definitions}
\centering
\small
\begin{tabularx}{\columnwidth}{@{}l X l@{}}
\toprule
\textbf{Outcome} & \textbf{Feasibility} & \textbf{Prediction} \\
\midrule
True Positive (TP)  & Infeasible state &  Infeasible prediction. \\
True Negative (TN)  & Feasible state & Feasible prediction. \\
False Positive (FP) & Feasible state & Infeasible prediction. \\
False Negative (FN) & Infeasible state & Feasible prediction. \\
\bottomrule
\end{tabularx}
\end{table}

\begin{figure}[t]
    \centering
    %\subcaptionbox{Confusion matrix\label{fig:confusion_matrix_total}}
        {\includegraphics[width=0.6\linewidth]{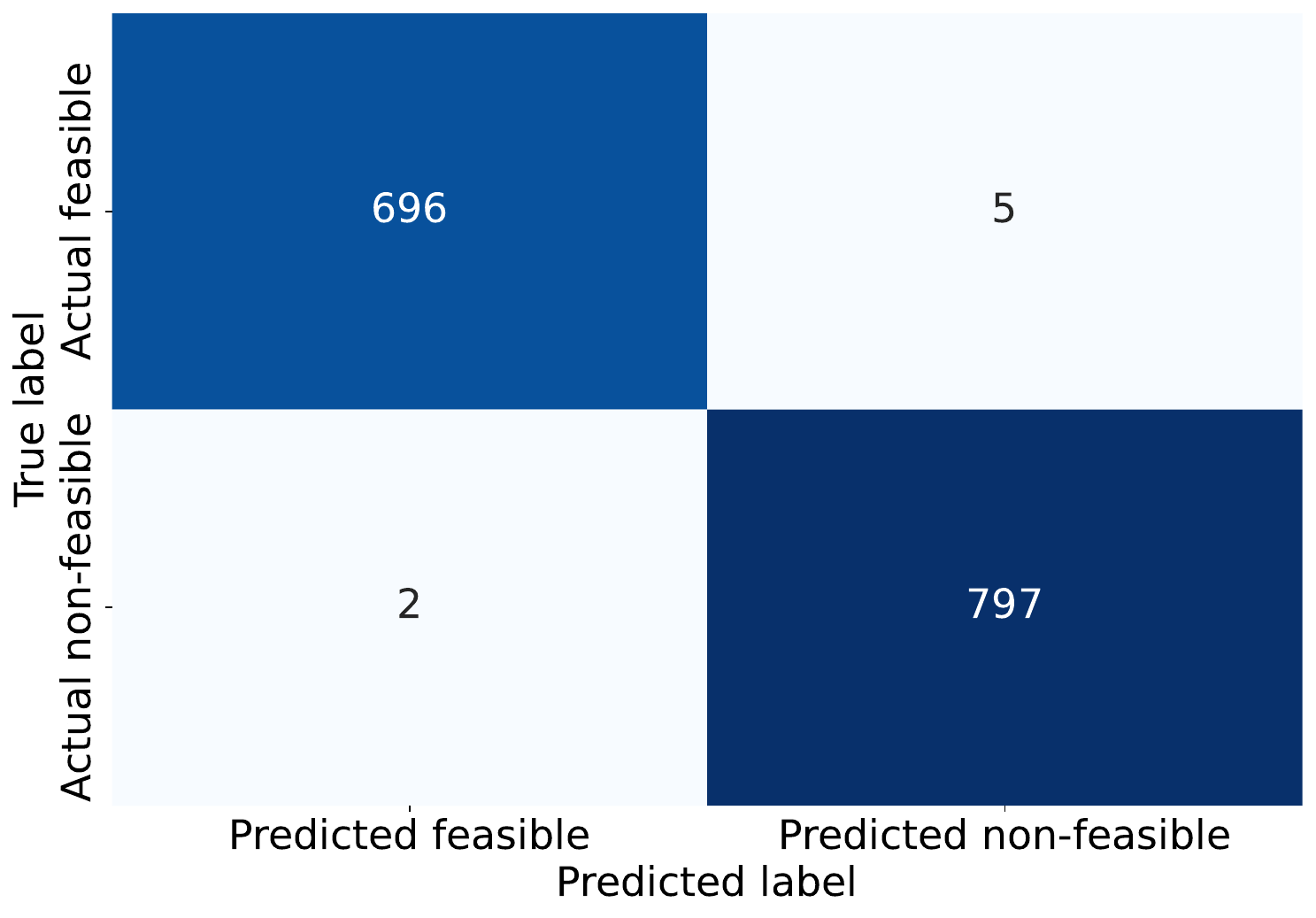}}
    %\hfill
    %\subcaptionbox{histogram of the graph reconstruction error.\label{fig:histogram_total}}
    %    {\includegraphics[width=0.49\linewidth]{images/histogram_total.png}}
    \caption{Combined dataset confusion matrix}
    \label{fig:after_training}
\end{figure}

The resulting detection performance is illustrated in Fig.~\ref{fig:after_training}, which shows the confusion matrix obtained after applying the threshold to the obtained reconstruction scores. The model correctly classifies most of the cases, with an overall accuracy of 99.5\%, with only 2 false negatives (see Table \ref{tab:confusion_definitions}) out of 1500 samples. Table ~\ref{tab:performance_metrics} details the results, done in each dataset and to the combination of them, after being trained with the Stressed Continuous train dataset, which is the most challenging setting, since multiple independent causes of infeasibility are represented, yielding a more general and robust model. Training with the Stressed Separated dataset, which introduces a bias toward power-only infeasibility, leads to significantly degraded generalization, with the model defaulting to predicting infeasibility for most feasible configurations.

To further analyze the obtained results, Fig.~\ref{fig:histogram_total_posttraining} shows the distribution of graph reconstruction errors for both classes: regular and anomalous. Feasible configurations are concentrated in the low-error region, whereas non-feasible configurations tend to exhibit higher reconstruction errors. The decision threshold is indicated by a vertical dashed line, which separates samples classified as regular from those classified as anomalous.

\begin{table}[!t]
\renewcommand{\arraystretch}{1.3}
\caption{Detection metrics by test dataset}
\label{tab:performance_metrics}
\centering
\small
% @{} removes extra side padding to save space
\begin{tabularx}{\columnwidth}{@{}l ccccc c@{}}
\toprule
\textbf{Scenario} & \textbf{\%Acc.}  & \textbf{TP} & \textbf{TN} & \textbf{FP} & \textbf{FN} \\
\midrule
SC trained SC dataset & 99.6 &  249 & 249 & 1 & 1 \\
SC trained RC dataset  & 99.2 &  250 & 246 & 4 & 0 \\
SC trained SS dataset  & 99.8 &  298 & 201 & 0 & 1 \\
SC trained combined dataset      & 99.5  & 797 & 696 & 5 & 2 \\
SS trained combined dataset        & 51.4  & 79 & 692 & 9 & 720 \\
\bottomrule
\end{tabularx}
\end{table}

\section{Conclusions}\label{sec:Conclusions}

This paper has demonstrated that the use of a~\gls{vgae}, based on a~\gls{mpnn} architecture, is an effective approach for detecting power flow feasibility. By training the model in an unsupervised manner exclusively on physically valid network states, the proposed framework successfully identifies infeasible configurations as anomalies characterized by high reconstruction errors.

The experimental results on the IEEE 118-bus system indicate that this method achieves a detection accuracy of 99.5\%. Critically, the model demonstrates high reliability for operational safety checks by maintaining a low rate of false negatives, which is the most significant risk in feasibility assessment. Furthermore, the study highlights the vital role of dataset diversity; it was shown that a model trained only on simple loading variations fails to generalize to complex infeasibility scenarios, whereas our multi-parametric approach, incorporating topological changes, operational limits, and stressed loading, enables robust learning.

The main contributions of this paper are:
\begin{itemize}
    \item An \textbf{unsupervised feasibility classifier} based on~\gls{vgae} that assesses network states without requiring labels from traditional iterative solvers.
    \item A \textbf{methodology for dataset generation} that captures diverse causes of infeasibility, including topological removals and operational limit violations, preventing the model from relying on trivial heuristics.
\end{itemize}

\printbibliography

@inproceedings{Trias2012,
  author    = {A. Trias},
  booktitle = {Proceedings of the IEEE PES General Meeting},
  title     = {The holomorphic embedding load flow method},
  year      = {2012}
}

@inproceedings{he2016deep,
  title={Deep residual learning for image recognition},
  author={He, Kaiming and Zhang, Xiangyu and Ren, Shaoqing and Sun, Jian},
  booktitle={Proceedings of the IEEE conference on computer vision and pattern recognition},
  pages={770--778},
  year={2016}
}

@article{kipf2016variational,
  title={Variational graph auto-encoders},
  author={Kipf, Thomas N and Welling, Max},
  journal={arXiv preprint arXiv:1611.07308},
  year={2016}
}

@article{li_convergence_2024,
	title = {Convergence criterion of power flow calculation based on graph neural network},
	volume = {2703},
	issn = {1742-6596},
	doi = {10.1088/1742-6596/2703/1/012042},
	language = {en},
	number = {1},
	journal = {Journal of Physics: Conference Series},
	publisher = {IOP Publishing},
	author = {Li, Shenhao and Pan, Zhichon and Li, Hongyi and Xiao, Yue and Liu, Ming and Wang, Xiaorui},
	month = feb,
	year = {2024},
	pages = {012042},

}

@article{scarselli2008graph,
  title={The graph neural network model},
  author={Scarselli, Franco and Gori, Marco and Tsoi, Ah Chung and Hagenbuchner, Markus and Monfardini, Gabriele},
  journal={IEEE transactions on neural networks},
  volume={20},
  number={1},
  pages={61--80},
  year={2008},
  publisher={IEEE}
}

@inproceedings{gilmer2017neural,
  title={Neural message passing for quantum chemistry},
  author={Gilmer, Justin and Schoenholz, Samuel S and Riley, Patrick F and Vinyals, Oriol and Dahl, George E},
  booktitle={International conference on machine learning},
  pages={1263--1272},
  year={2017},
  organization={Pmlr}
}

@article{tinney_power_1967,
	title = {Power {Flow} {Solution} by {Newton}'s {Method}},
	volume = {PAS-86},
	issn = {0018-9510},
	doi = {10.1109/TPAS.1967.291823},
	number = {11},
	journal = {IEEE Transactions on Power Apparatus and Systems},
	author = {Tinney, William. F. and Hart, Clifford E.},
	month = nov,
	year = {1967},
	pages = {1449--1460},
}

@inproceedings{klump_techniques_2000,
	title = {Techniques for improving power flow convergence},
	volume = {1},
	doi = {10.1109/PESS.2000.867654},
	booktitle = {2000 {Power} {Engineering} {Society} {Summer} {Meeting} ({Cat}. {No}.{00CH37134})},
	author = {Klump, R.P. and Overbye, T.J.},
	month = jul,
	year = {2000},
	pages = {598--603 vol. 1},
}

@article{fatah_integrating_2021,
	title = {Integrating power grid topology in graph neural networks for power flow},
	language = {en},
	author = {Fatah, Mukhlish Ghany Al},
	month = mar,
	year = {2021},
}

@article{ajjarapu_continuation_1992,
	title = {The continuation power flow: a tool for steady state voltage stability analysis},
	volume = {7},
	issn = {1558-0679},
	shorttitle = {The continuation power flow},
	doi = {10.1109/59.141737},
	number = {1},
	journal = {IEEE Transactions on Power Systems},
	author = {Ajjarapu, V. and Christy, C.},
	month = feb,
	year = {1992},
	pages = {416--423},
}

@article{donon_neural_2020,
	title = {Neural networks for power flow: {Graph} neural solver},
	volume = {189},
	issn = {0378-7796},
	shorttitle = {Neural networks for power flow},
	doi = {10.1016/j.epsr.2020.106547},
	journal = {Electric Power Systems Research},
	author = {Donon, Balthazar and Clément, Rémy and Donnot, Benjamin and Marot, Antoine and Guyon, Isabelle and Schoenauer, Marc},
	month = dec,
	year = {2020},
	pages = {106547},
}

\end{document}